\documentclass[a4paper,conference]{IEEEtran}
\usepackage{xcolor}

\usepackage[T1]{fontenc}
\usepackage[latin9]{inputenc}
\usepackage{array}
\usepackage{units}
\usepackage{multirow}
\usepackage[pdftex]{graphicx} 
\usepackage{subcaption}
\usepackage[cmex10]{amsmath}
\usepackage{amssymb}
\usepackage{amsfonts}
\usepackage{amsmath}
\usepackage{supertabular}
\usepackage{times}
\usepackage{amsthm}
\usepackage{floatrow}
\usepackage{tikz}
\usepackage{float}
\usepackage{mathtools}
\usepackage{textcomp}
\usepackage{algpseudocode}

\usepackage{adjustbox}
\usepackage[justification=centering, font=small]{caption} 
\usepackage{epstopdf}
\usepackage{tabularx}
\usepackage{nccmath}
\usepackage{listings}
\usepackage{mathtools}

\usepackage{hyperref}

\usepackage[noline,ruled,algo2e,linesnumbered]{algorithm2e}

\usepackage{microtype}
\usepackage{graphicx}

\usepackage{booktabs} 
\usepackage{comment}
\usepackage{amsmath,amssymb,amsfonts}

\usepackage{xcolor}
\usepackage{multirow}
\usepackage{array}
\usepackage{units}
\usepackage{wrapfig}
\usepackage{amsthm}
\usepackage{nccmath}
\usepackage{mathtools}
\usepackage{textcomp}
\usepackage{adjustbox}
\usepackage{epstopdf}
\usepackage{tabularx}
\usepackage{subcaption}

\newcommand{\isabelle}[1]{\textcolor{orange}{Isabelle: #1}}

\newcommand{\PB}[1]{\textcolor{red}{Prasanna: #1}}

\DeclareMathOperator*{\argmax}{arg\,max}
\DeclareMathOperator*{\argmin}{arg\,min}

\title{AutoDEUQ: Automated Deep Ensemble with Uncertainty Quantification}

\makeatletter
\newcommand{\linebreakand}{%
  \end{@IEEEauthorhalign}
  \hfill\mbox{}\par
  \mbox{}\hfill\begin{@IEEEauthorhalign}
}
\makeatother

\author{\IEEEauthorblockN{Romain Egele}
\IEEEauthorblockA{Argonne National Laboratory, USA\\
\& Universit\'e Paris-Saclay, France\\
romain.egele@universite-paris-saclay.fr}
\and
\IEEEauthorblockN{Romit Maulik}
\IEEEauthorblockA{Argonne National Laboratory\\
Lemont, Illinois, USA\\
rmaulik@anl.gov}
\and
\IEEEauthorblockN{Krishnan Raghavan}
\IEEEauthorblockA{Argonne National Laboratory\\
Lemont, Illinois, USA\\
kraghavan@anl.gov}
\linebreakand 
\IEEEauthorblockN{Bethany Lusch}
\IEEEauthorblockA{Argonne National Laboratory\\
Lemont, Illinois, USA\\
blusch@anl.gov}
\and
\IEEEauthorblockN{Isabelle Guyon}
\IEEEauthorblockA{Universit\'e Paris-Saclay\\
Paris, France\\
isabelle.guyon@universite-paris-saclay.fr}
\and
\IEEEauthorblockN{Prasanna Balaprakash}
\IEEEauthorblockA{Argonne National Laboratory\\
Lemont, Illinois, USA\\
pbalapra@anl.gov}
}
\begin{document}
\maketitle

\begin{abstract}
Deep neural networks are powerful predictors for a variety of tasks. However, they do not capture uncertainty directly. Using neural network ensembles to quantify uncertainty is competitive with approaches based on Bayesian neural networks while benefiting from better computational scalability. However, building ensembles of neural networks is a challenging task because, in addition to choosing the right neural architecture or hyperparameters for each member of the ensemble, there is an added cost of training each model. To address this issue, we propose AutoDEUQ, an automated approach for generating an ensemble of deep neural networks. Our approach leverages joint neural architecture and hyperparameter search to generate ensembles. We use the law of total variance to decompose the  predictive variance of deep ensembles into aleatoric (data) and epistemic (model) uncertainties. We show that AutoDEUQ outperforms probabilistic  backpropagation,  Monte  Carlo  dropout, deep ensemble, distribution-free ensembles, and hyper ensemble methods on a number of regression benchmarks. 
\end{abstract}

\section{Introduction}


Uncertainty quantification (UQ) for machine-learning-based predictive models is crucial for assessing the trustworthiness of predictions from the trained model. For deep neural networks (DNNs), it is desirable for predictions to be accompanied with estimates of uncertainty because of the black-box nature of the function approximation. Two major forms of uncertainty  exist~\cite{hullermeier2021aleatoric}: aleatoric data uncertainty and epistemic model uncertainty. The former occurs due to the inherent variability or noise in the data. The latter is attributed to the uncertainty associated with the NN model parameter estimation or out-of-distribution predictions. The epistemic uncertainty increases in the regions that are not well represented in the training dataset~\cite{gal2016dropout}. While the aleatoric uncertainty is irreducible, the epistemic uncertainty can be reduced by collecting more training data in the appropriate regions.   

Several  researchers  have looked at extending deterministic neural networks to probabilistic models. A strongly advocated method is to have a fully Bayesian formulation, where each trainable parameter in a DNN is assumed to be sampled from a very high-dimensional (and arbitrary) joint distribution~\cite{neal2012bayesian}. However, this is computationally infeasible, for example because of issues of convergence, for any practical deep learning tasks with millions of trainable parameters in the architecture and having large datasets. Consequently, several approximations to fully Bayesian formulations have been put forth to reduce the computational complexity of uncertainty quantification in DNNs. These range from simple augmentations such as the mean-field approximation in Bayesian backpropagation via variational inference~\cite{hoffman2013stochastic,hernandez-lobato_probabilistic_2015}, where each parameter is assumed to be sampled from an independent unimodal Gaussian distribution, to Monte Carlo dropout~\cite{srivastava2014dropout}, where random neurons are switched off during training and inference to obtain ensemble predictions. 

Ensemble methods that utilize multiple independently trained DNNs have shown considerable promise for uncertainty quantification~\cite{lakshminarayanan_simple_2017,ovadia2019can,ashukha2020pitfalls} by outperforming conventional approximations to the fully Bayesian methodology. Blundell et al.~\cite{wilson2020bayesian} argue that the deep ensembles approach is fully congruous with Bayesian model averaging, which attempts to estimate the posterior distribution of the targets given input data by marginalizing the parameters. However, a key factor in deep ensembles is model diversity 
without which uncertainty cannot be captured efficiently. For example, in \cite{lakshminarayanan_simple_2017},  each member of the ensemble has an identical neural architecture and is trained by using maximum likelihood or maximum a posteriori optimization through different initialization of weights. Consequently, ensemble diversity is limited since each model can at best settle on distinct local minima. Marginalization over these models in the ensemble will force the function approximation to collapse on one hypothesis and provide results similar to Bayesian model averaging for a single architecture with probabilistic trainable parameters.  Such an implicit assumption may be undesirable when dealing with datasets that are generated from a combination of hypotheses. Moreover, the lack of flexibility in the ensemble may lead to a poorer estimate of epistemic uncertainty. Although Wenzel et al.~\cite{wenzel_hyperparameter_2021} attempted to relax this issue by allowing more diversity in the ensemble, they vary just two hyperparameters. Similarly, Zaidi et al.~\cite{zaidi2020neural} vary the architecture with fixed trainable hyperparameters to increase the ensemble diversity. By constructing diverse DNNs models through a methodical and automated approach, we hypothesize that the assumption of and the eventual collapse to one hypothesis can be avoided, thus providing robust and efficient estimates of uncertainty. 




\subsection{Related Work}
To model aleatoric uncertainty, one must model the conditional distribution $p(y \mid \mathbf{x})$ for the target $y$ given an input $\textbf{x}$. One way is to assume that this distribution is Gaussian and then estimate its parameters~(mean and variance)~\cite{374138}. However, these estimates summarize conditional distributions into scalar values and are thus unable to model more complex profiles of uncertainty such as multimodal or heteroscedastic profiles. To resolve this issue, one can use implicit generative models~\cite{mohamed2016learning} and mixture density networks~\cite{bishop1994mixture}. A different approach is deep kernel learning~\cite{van2021feature}, which extracts kernels and uses them in Gaussian-process-based methods for datasets with large features and sample size. However, this adds additional complexity because one must find the correct hyperparameters. An alternative strategy is to directly output prediction intervals from the NN, such as in~\cite{pearce2018high}, which has the advantage of not requiring any distribution assumption on the output variables. However, these methods are ill-equipped to quantify epistemic uncertainty.

Several methods for epistemic uncertainty have been proposed. Bayesian NNs (BNNs)~\cite{maddox_simple_nodate} and deep ensembles~\cite{caruana_getting_2006} are the main approaches. In BNN, the weights are assumed to follow a joint distribution, and the epistemic uncertainty is quantified through Bayesian inference. Except for  trivial cases, however, Bayesian inference is computationally intractable. Therefore, several approximations to BNN have been proposed, such as probabilistic backpropagation (PBP)~\cite{hernandez-lobato_probabilistic_2015} and Bayes by Backprop~\cite{blundell2015weight}. 
In deep ensembles~\cite{lakshminarayanan_simple_2017}, multiple networks are aggregated to quantify the uncertainty. 
Each network in the ensemble provides an estimate of aleatoric uncertainty, while their aggregation provides an estimate of epistemic uncertainty. However, the members of such ensembles often have similar architecture and hyperparameter values but with different weights generated through random weight initialization in addition to the stochastic aspect of the training procedure. Recently, new automated methods were proposed to improve deep ensembles, wherein  hyperparameters~\cite{wenzel_hyperparameter_2021} or neural architecture decision variables~\cite{zaidi2020neural} are varied to improve the diversity of models in the ensemble to achieve improved aleatoric and epistemic uncertainty estimates. 

Recently, Russell and Reale \cite{russell2021multivariate} developed a joint covariance matrix with end-to-end training using a Kalman filter to represent aleatoric uncertainty while using dropout to estimate the epistemic component. Although not an ensemble method, it models aleatoric and epistemic at the same time.

\subsection{Contributions}
Given training and validation data, the proposed AutoDEUQ method (i) starts from a user-defined neural architecture and hyperparameter search space; (ii) leverages aging evolution and Bayesian optimization methods to automatically tune the architecture decision variables and training hyperparameters, respectively; (iii) builds a catalog of models from the search; and (iv) uses a greedy heuristic to select models from the catalog to construct ensembles. The predictions from the ensemble models are then used to estimate the aleatoric and epistemic uncertainty. 
AutoDEUQ is built on the successes of three recent works in the deep ensemble literature: deep ensemble~\cite{lakshminarayanan_simple_2017}, hyper ensemble~\cite{wenzel_hyperparameter_2021}, and neural ensemble search \cite{zaidi2020neural}. However, our AutoDEUQ method differs from deep ensemble in the following ways. While aleatoric and epistemic uncertainties are modeled empirically, we theoretically decompose the predicted variance of deep ensembles into its aleatoric and epistemic components. Moreover, in AutoDEUQ, the DNN architectures and the training hyperparameter values in the ensembles are different, and more importantly they are generated automatically. While hyper ensemble and neural ensemble methods explore hyperparameters and architectural choices, respectively, and generate ensembles, AutoDEUQ explores both spaces simultaneously. 
The key contributions of the paper are as  follows: (1) automation of deep ensembles construction with joint neural architecture and hyperparameter search and (2) demonstration of improved uncertainty quantification compared with prior ensemble methods and, consequently, advancement of state of the art in deep ensembles.

\section{AutoDEUQ}

We focus on uncertainty estimation in a regression setting. Our methodology, automated deep ensemble for uncertainty quantification (AutoDEUQ), estimates aleatoric and epistemic uncertainties by
automatically generating a catalog of NN models through joint neural architecture and hyperparameter search, wherein each model is trained to minimize the negative log likelihood to capture aleatoric uncertainty, and
selecting a set of models from the catalog to construct the ensembles and model epistemic uncertainty without losing the quality of aleatoric uncertainty.



In supervised learning, the dataset $\mathcal{D}$ is composed of i.i.d points $\left(\mathbf{x}_{i} \in \mathcal{X}, y_{i} \in \mathcal{Y}\right)$, where $\mathbf{x}_{i}$ and $y_{i}=f\left(\mathbf{x}_{i}\right)$ are the input and the corresponding output of the $i$th point, respectively, and $\mathcal{X} \subset \mathbb{R}^N$ and $\mathcal{Y} \subset \mathbb{R}^M$ are the input and output spaces of $N$ and $M$ dimensions, respectively. Here, we focus on regression problems, wherein the output is a scalar or vector of real values. Given $\mathcal{D}$, we seek to model the probabilistic predictive distribution $p(y|\mathbf{x})$ using a parameterized distribution $p_\theta(y|\mathbf{x}),$ which estimates aleatoric uncertainty through a trained NN and then estimates the epistemic uncertainty with an ensemble of NNs $p_\mathcal{E}(y|\mathbf{x})$. We define $\Theta$ to be the sample space for $\theta.$

The aleatoric uncertainty can be modeled by using the quantiles of $p_{\theta}.$ Following previous work~\cite{lakshminarayanan_simple_2017}, we assume a Gaussian distribution for $p_{\theta} \sim \mathcal{N}(\mu_\theta, \sigma_\theta^2)$ and use variance as a measure of the aleatoric uncertainty. We explicitly partition $\theta$ into $(\theta_a, \theta_h, \theta_w)$ such that $\Theta$ is decomposed into $(\Theta_a, \Theta_h, \Theta_w)$, where $\theta_a \in \Theta_a$ represents the NN values of the architecture decision variables (network topology parameters), $\theta_h \in \Theta_h$ represents  NN training hyperparameters (e.g., learning rate, batch size), and $\theta_w \in \Theta_w$ represents the NN weights. The NN is trained to output mean $\mu_{\theta}$ and variance $\sigma_{\theta}^2$. For a given choice of architecture decision variables $\theta_a$ and training hyperparameters $\theta_h$, to obtain $\theta_w^*,$ we seek to maximise the likelihood given the real data $\mathcal{D}$. Specifically, we can model the aleatoric uncertainty using the negative log-likelihood loss (as opposed to the usual mean squared error) in the training: ~\cite{lakshminarayanan_simple_2017}:
\begin{align}
    \label{eq:ll_func}
    \ell(\mathbf{x}, y; \theta) = -\log p_\theta =\frac{\log \sigma_\theta^{2}(\mathbf{x})}{2}+\frac{\left(y-\mu_{\theta}(\mathbf{x})\right)^{2}}{2 \sigma_{\theta}^{2}(\mathbf{x})}+ \text{cst},
\end{align}
where cst is a constant. The NN training problem is then 
\begin{equation} \label{eq:opt}
    \theta_w^* = \argmax_{\theta_w \in \Theta_w} \quad \ell(\mathbf{x}, y; \theta_a, \theta_h, \theta_w).
\end{equation}

To model epistemic uncertainty, we use deep ensembles~(an ensemble composed of NNs) \cite{lakshminarayanan_simple_2017}. In our approach, we generate a catalog of NN models $\mathcal{C} = \{\theta_i, i = 1, 2, \cdots, c\}$ (where $\theta \in \Theta$ is a tuple of architecture, optimization hyperparameters, and weights) and repeatedly sample $K$ models to form the ensemble $\mathcal{E} | \mathcal{E} =\{ \theta_i, i=1,2,\cdots,K \}$. 
Let $p_{\theta}$ describe the probability that $\theta$ is a member of the ensemble $\forall \theta \in \mathcal{C}.$ Let $p_\mathcal{E}$---the probability density function of the ensemble---be obtained as a mixture distribution where the mixture is given as  $ p_\mathcal{E} = \mathbb{E} p_{\theta}.$
Define $\mu_{\theta}$ and $\sigma_{\theta}^2$ as the mean and variance of each element in the ensemble, respectively. Then,  the mean of the mixture is $\mu_{\mathcal{E}} := \mathbb{E}[\mu_{\theta}]$, and the variance~\cite{rudary2009predictive} is
\begin{equation} \label{eq:variance-mixture-regression_prop}
\begin{split}
\sigma_{\mathcal{E}}^2 &:= \mathbb{V}[p_\mathcal{E}] = \underbrace{\mathbb{E} [\sigma_{\theta}^2]}_{ \text{Aleatoric Uncertainty} } + \underbrace{ \mathbb{V}[\mu_{\theta}]}_{\text{Epistemic Uncertainty} }, \\
\end{split}
\end{equation}
where $\mathbb{E}$ refers to the expected value and $\mathbb{V}$ refers to the variance. Equation \eqref{eq:variance-mixture-regression_prop} formally provides the decomposition of overall uncertainty of the ensemble into its individual components such that $ \mathbb{E} [\sigma_{\theta}^2]$ marginalizes the effect of $\theta$ and captures the aleatoric uncertainty and $\mathbb{V}[\mu_{\theta}]$ captures the spread of the prediction across different models and neglects the noise of the data, therefore capturing the epistemic uncertainty.

We write the empirical estimate of the mean and variance  as
\begin{equation}
\begin{split}
\mu_{\mathcal{E}}     &  = \frac{1}{K}\sum_{\theta \in \mathcal{E} } \mu_{\theta} \\
\sigma_\mathcal{E}^2  &  = \underbrace{\frac{1}{K} \sum_{\theta \in \mathcal{E}} \sigma^2_\theta}_{ \text{Aleatoric Uncertainty} } + \underbrace{\frac{1}{K-1} \sum_{\theta \in \mathcal{E}} (\mu_\theta - \mu_\mathcal{E})^2}_{ \text{Epistemic Uncertainty}},
\label{eq:var-decomposition-2}
\end{split}
\end{equation}
where $K$ is the size of the ensemble. The total uncertainty quantified by $\sigma_\mathcal{E}^2 $ is a combination of aleatoric and  epistemic uncertainty, which are given by the the mean of the predictive variance of each model in the ensemble and the predictive variance of the mean of each model in the ensemble. 


\subsection*{Catalogue generation and ensemble construction}

Let $\mathcal{D}$ be decomposed as $\mathcal{D} = \mathcal{D}^{train} \cup \mathcal{D}^{valid} \cup \mathcal{D}^{test}$, referring to the training, validation, and test data, respectively. A neural architecture configuration $\theta_a$ is a vector from the neural architecture search space $\Theta_a$, defined by a set of neural architecture decision variables. A hyperparameter configuration $\theta_h$ is a vector from the training hyperparameter search space $\Theta_h$ defined by a set of hyperparameters used for training (e.g., learning rate, batch size). The problem of joint neural architecture and hyperparameter search can be formulated as the following bilevel optimization problem:  
\begin{equation}
    \begin{aligned}
        &\theta_a^*, \theta_h^* = \argmax_{\theta_a,\theta_h} \frac{1}{N^{valid}}\sum_{\mathbf{x}, y \in \mathcal{D}^{valid}}\ell(\mathbf{x}, y;\theta_a,\theta_h,\theta_w^*) \\
        & \text{s.t. }  \theta_w^* =  \argmax_{\theta_w} \frac{1}{N^{train}}\sum_{\mathbf{x}, y \in \mathcal{D}^{train}} \ell(\mathbf{x}, y; \theta_a, \theta_h, \theta_w),
    \end{aligned}
    \label{eqn:pbopti3}
\end{equation}
where the best architecture decision variables $\theta_a^*$ and training hyperparameters values  $\theta_h^*$ are selected based on the validation set and the corresponding weights $\theta_w$ are selected based on the training set.


The pseudo code of the AutoDEUQ is shown in the Appendix, Algorithm \ref{alg:AgEBO}. To perform a joint neural architecture and hyperparameter search, we leverage aging evolution with asynchronous Bayesian optimization (AgEBO)~\cite{egele2020agebo}. 
Aaging evolution (AgE)~\cite{real_regularized_2018} is a parallel neural architecture search (NAS) method for searching over the architecture space.
The AgEBO method follows the manager-worker paradigm, wherein a manager node runs a search method to generate multiple NNs and $W$ workers (compute nodes) train them simultaneously. The AgEBO method constructs the initial population by sampling $W$ architecture and  $W$ hyperparameter configurations and concatenating them (lines~1--7). The NNs obtained by using these concatenated configurations are sent for simultaneous  evaluation on $W$ workers (line~6). The iterative part (lines~8--26) of the method checks whether any of the workers finish their evaluation (line~9), collects validation metric values from the finished workers, and uses them to generate the next set of architecture and hyperparameter configurations for simultaneous evaluation to fill up the free workers that finished their evaluations (lines~11--25). 
At a given iteration, in order to generate a NN, architecture and hyperparameter configurations are generated in the following way. From the incumbent population, $S$ NNs are sampled (line~17). A random mutation is applied to the best of $S$ NNs to generate a child architecture configuration (line~18). This mutation is obtained by first randomly selecting an architecture decision variable from the selected NN and replacing its value with another randomly selected value excluding the current value. The new child replaces the oldest member of the population. The AgEBO optimizes the hyperparameters ($\theta_h$) by marginalizing the architecture decision variables ($\theta_a$). At a given iteration, to generate a hyperparameter configuration, the AgEBO uses a (supervised learning) model $M$ to predict a point estimate (mean value) $\mu(\theta_h^i)$ and standard deviation $\sigma(\theta_h^i)$ for a large number of unevaluated hyperparameter configurations. The best configuration is selected by ranking all sampled hyperparameter configurations using the upper confidence bound  acquisition function, which is parameterized by $\kappa \geq 0$ that controls the trade-off between exploration and exploitation. To generate multiple hyperparameter configurations at the same time, the AgEBO leverages a multipoint acquisition function based on a constant liar strategy~\cite{hiot_kriging_2010}. 

The catalog $\mathcal{C}$ of NN models is obtained by running AgEBO and storing all the models from the runs. To build the ensemble  $\mathcal{E}$ of models from $\mathcal{C}$, we adopt a greedy selection strategy (lines~27--38) ~\cite{caruana_ensemble_2004}.  
At each step, the model from the catalog that most improves the negative log likelihood of the incumbent ensemble is added to the ensemble. The greedy approach can work well when the validation data is representative of the generalisation task (i.e., big enough, diverse enough, with good coverage)~\cite{caruana_ensemble_2004}. 


\section{Results}

We first describe the search space used in AutoDEUQ. Next, using a one-dimensional dataset, we present an ablation study to analyze the impact of different components of AutoDEUQ. Then, we compare AutoDEUQ with other methods. 


\subsection{Search Space}

The architecture search space is modeled by using a directed acyclic graph, which starts and ends with input and output nodes, respectively (see Appendix for an illustration). They represent the input and output layer of NN, respectively. Between the two are intermediate nodes defined by a series of variable $\mathcal{N}$ and skip connection $\mathcal{SC}$ nodes. Both types of nodes correspond to categorical decision variables.
The variable nodes model dense layers with a list of different layer configurations. The skip connection node creates a skip connection between the variable nodes. This second type of node can take two values: disable or create the skip connection.
For a given pair of consecutive variable nodes $\mathcal{N}_{k}$, $\mathcal{N}_{k+1}$, three skip connection nodes $\mathcal{SC}^{k+1}_{k-3}, \mathcal{SC}^{k+1}_{k-2}, \mathcal{SC}^{k+1}_{k-1}$  are created. These  nodes allow for connection to the previous nonconsecutive variable nodes $\mathcal{N}_{k-3}, \mathcal{N}_{k-2}, \mathcal{N}_{k-1}$, respectively. 
Each dense layer configuration is defined by the number of units and the activation function. We used values in \{16, 32, ..., 256\}  and \{elu, gelu, hard sigmoid, linear (i.e., identity), relu, selu, sigmoid, softplus, softsign, swish, and tanh\}, respectively. These resulted in 177 (16 units $\times$ 11 activation functions, and identity) dense layer types for each variable node. Skip connections can be created from at most 3 previous dense layers. Each skip connection is created with a linear projection so that feature vectors match in shape, and then addition is used to merge the vectors. The number of variable nodes is set to 3 for the one-dimensional toy dataset and to 5 for the regression benchmarks.

For the hyperparameter search space, we use a learning rate in the continuous range $[10^{-4},10^{-1}]$ with a log-uniform prior;  a batch size in the discrete range $[1,2,3,\ldots,b_{max}]$ (where $b_{max}=32$ for the toy example and $b_{max}=256$ for the benchmark) with a log-uniform prior; an optimizer in $\{\text{sgd}, \text{rmsprop}, \text{adagrad}, \text{adam}, \text{adadelta}, \text{adamax}, and \text{nadam}\}$; a patience number  for the reduction of the learning rate in the discrete range $[10,11,\ldots,20]$, and a patience number for early stopping in the discrete range $[20,21,\ldots,30]$. The NNs are trained with 200 epochs for the toy example and 100 epochs for the benchmark. The search space is the same for the toy and the benchmark. Models are checkpointed during their evaluation based on the minimum validation loss achieved. Input and output variables are standardized to have a mean of 0 and a unit variance.

The hardware and software platforms as well as other execution settings are described in the Appendix.

\subsection{Toy Example}

We follow the ideas from ~\cite{hernandez-lobato_probabilistic_2015} to assess qualitatively the effectiveness of AutoDEUQ on a one-dimensional dataset. However, instead of the unimodal dataset generated from the cubic function used in ~\cite{hernandez-lobato_probabilistic_2015}, we used the $y = f(x) = 2 \sin{x} + \epsilon$ sine function. We generated 200 points randomly sampled from a uniform prior in the x-range $\left[-30,-20\right]$ with $\epsilon \sim \mathcal{N}(0,0.25)$ and 200 other points randomly sampled in the x-range $\left[20,30\right]$ with $\epsilon \sim \mathcal{N}(0,1)$. These 400 points constitute $\mathcal{D}^{train} \cup \mathcal{D}^{valid}$. We used random sampling to split the generated data: 2/3 for training and 1/3 for validation datasets. The two x-ranges are sampled with different noise levels to assess the learning of aleatoric uncertainty. The test set comprised 200 x-values regularly spaced between $[-40,40]$, and the corresponding y values were given by $2 \sin{x}$ with $\epsilon = 0$. Consequently, we had three different ranges of x-values to assess epistemic uncertainty: training region, $\left[-30,-20\right]$ and $\left[20,30\right]$; interpolation region, $\left[-20,20\right]$; and extrapolation region: $\left[-40,-30\right]$ and $\left[30,40\right]$. We seek to verify that the proposed method can model the aleatoric (different noise levels in the training region) and epistemic uncertainty (interpolation and extrapolation regions).

\subsubsection{Ablation study of catalog generation}

\begin{figure}
     \centering
     \includegraphics[width=\textwidth]{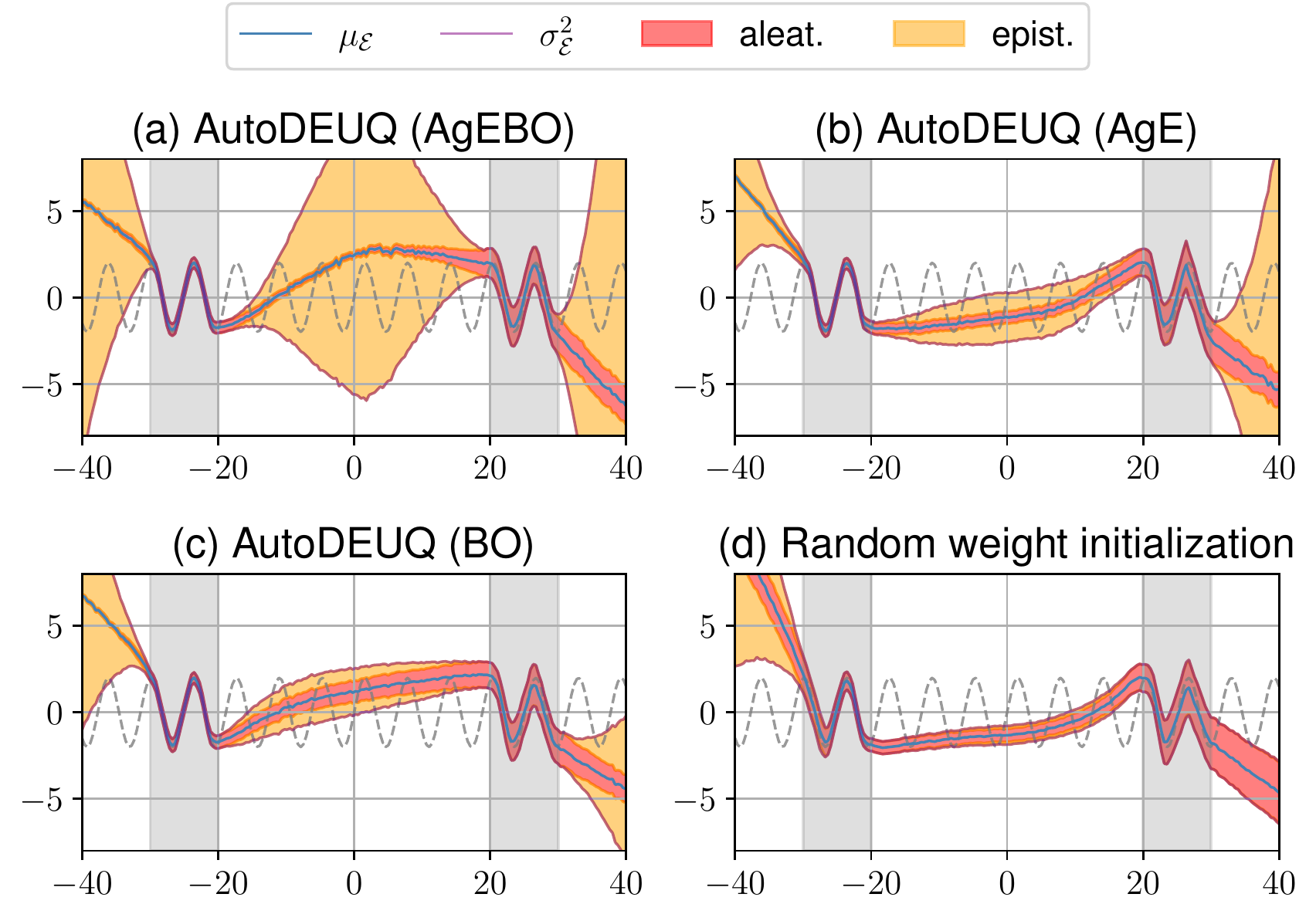}
    \caption{Ablation study of catalog generation: We progressively removed the different algorithmic components of AutoDEUQ and analyzed their impact on the uncertainty estimation.}
    \label{fig:ablation_study}
    \vspace{-0.5cm}
\end{figure}




We perform an ablation study to show the effectiveness of tuning both architecture decision variables and training hyperparameters in AutoDEUQ. First, we designed a high-performing NN by manually tuning the architecture decision variables and hyperparameter configurations on the validation data (see the Appendix for the obtained values).
We ran AutoDEUQ, which  used AgEBO for catalog generation and the greedy model selection method for ensemble construction. Next, we used two AutoDEUQ variants: (1) AutoDEUQ (AgE), which used only AgE  to explore the search space of the architecture space but used the hand-tuned hyperparameter values following the approach from~\cite{zaidi2020neural}, and (2) AutoDEUQ (BO), which used the hand-tuned neural architecture and used BO to tune the hyperparameters following the approach  from~\cite{wenzel_hyperparameter_2021}. Finally, we switched off both AgE and BO and trained the manually generated baseline with $500$ random-weight initializations to build the catalog. All these methods used greedy selection to build an ensemble of size $K=5$ from their respective catalog of 500 models.

Figure \ref{fig:ablation_study} shows the results of  these  variants. We  observe that the proposed AutoDEUQ (Fig.~2.a) obtains a superior aleatoric and epistemic uncertainty estimation. The two different noise levels in the training region are well captured by the aleatoric uncertainty estimate. In the interpolation region, aleatoric uncertainty follows the noise levels of the nearby region.
We also observe that epistemic uncertainty grows as we move from the training data region (grey). Moreover, we  observe that its magnitude is large for the extrapolation region compared with the interpolation regions. Unlike AutoDEUQ (AgE) and AutoDEUQ (BO),  the epistemic uncertainty grows from $x=-20$, peaks near $x=0$, and becomes zero near $x=20$. The results of AutoDEUQ (AgE) and AutoDEUQ (BO) variants are similar: while the aleatoric uncertainty estimates are good, both suffer from poor epistemic uncertainty estimation in the interpolation region. This can be attributed to a lack of model diversity in the ensemble, the former with fixed hyperparameters and the latter with fixed architectures. We observe that the random initialization strategy (Fig.~2.d) with the hand-tuned neural architecture did not model epistemic uncertainty well. This result can be attributed to the simplicity of the dataset: given its low dimension, for the same architecture and hyperparameter configurations, the training results in similar NN models.

\subsubsection{Comparison of search methods}


\begin{figure}
     \centering
     \includegraphics[width=\textwidth]{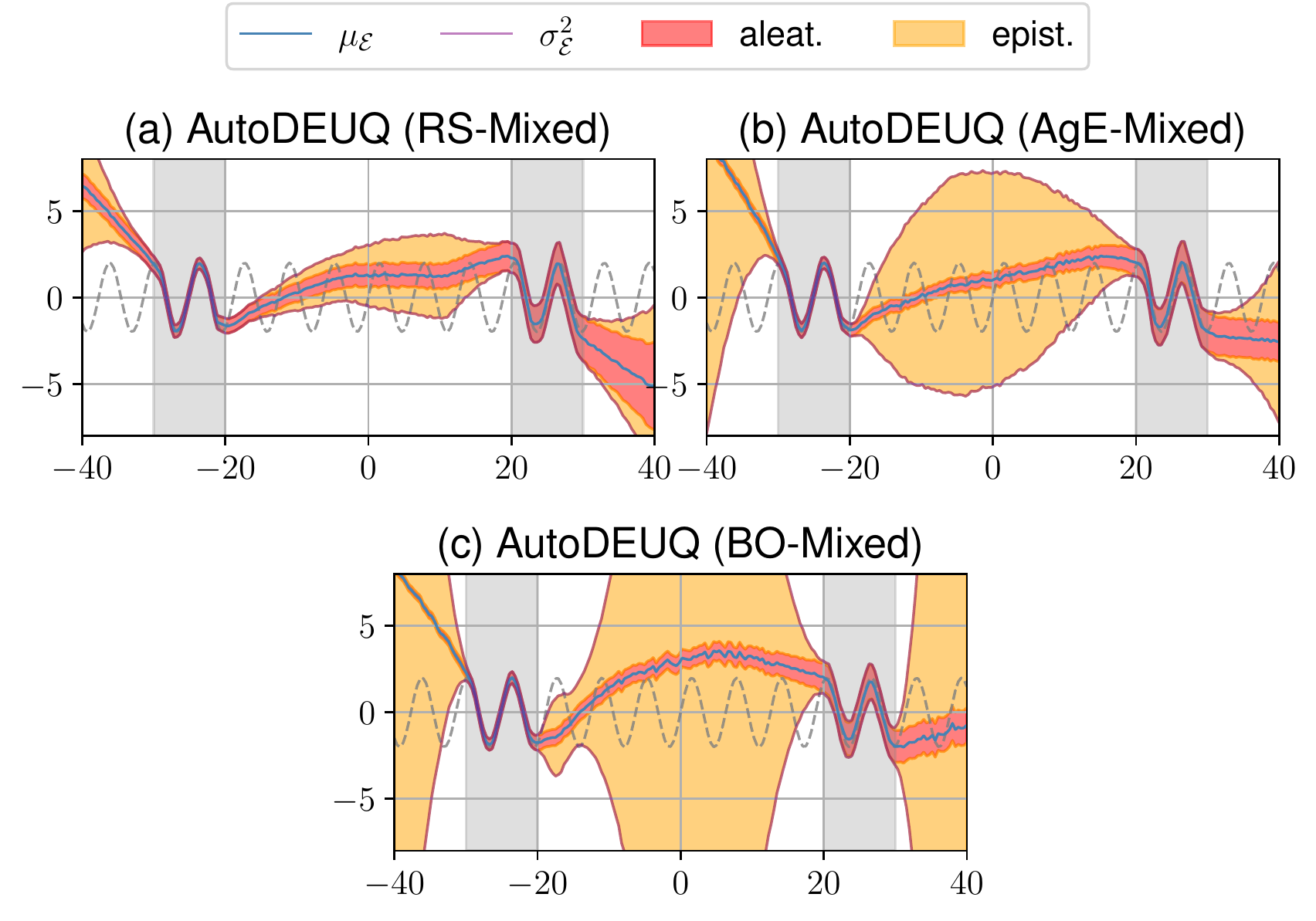}
    \caption{Comparison of different search methods in AutoDEUQ and their impact on uncertainty estimation.}
    \label{fig:different_search_for_catalog}
    \vspace{-0.5cm}
\end{figure}

\begin{table*}[h]
\centering
\resizebox{0.9\textwidth}{!}{%
\begin{tabular}{|c|cccccc|cccccc|}
\hline
\multirow{2}{*}{\textbf{Dataset}} & \multicolumn{6}{c|}{\textbf{NLL}}                                                                                                                                                                                                                                                          & \multicolumn{6}{c|}{\textbf{RMSE}}                                                                                                                                                                                                                                                          \\
                                  & PBP   & \begin{tabular}[c]{@{}c@{}}MC\\ Dropout\end{tabular} & \begin{tabular}[c]{@{}c@{}}Deep\\ Ens.\end{tabular} & \begin{tabular}[c]{@{}c@{}}Hyper\\ Ens.\end{tabular} & \begin{tabular}[c]{@{}c@{}}DF\\ Ens.\end{tabular} & \begin{tabular}[c]{@{}c@{}}AutoDEUQ \\ \end{tabular} & PBP  & \begin{tabular}[c]{@{}c@{}}MC\\ Dropout\end{tabular} & \begin{tabular}[c]{@{}c@{}}Deep \\ Ens.\end{tabular} & \begin{tabular}[c]{@{}c@{}}Hyper \\ Ens.\end{tabular} & \begin{tabular}[c]{@{}c@{}}DF\\ Ens.\end{tabular} & \begin{tabular}[c]{@{}c@{}}AutoDEUQ \\ \end{tabular} \\ \hline
boston                            & 2.57  & 2.46                                                 & 2.41                                                & \textbf{2.15 (0.22)}                                 & 2.74                                              & 2.46 (0.09)                                                & 3.01 & 2.97                                                 & 3.28                                                 & \textbf{2.87 (0.1)}                                   & 3.38                                              & 3.09 (0.31)                                                \\
concrete                          & 3.16  & 3.04                                                 & 3.06                                                & 4.09 (0.17)                                          & 3.10                                              & \textbf{2.86 (0.07)}                                       & 5.67 & 5.23                                                 & 6.03                                                 & 4.7 (0.08)                                            & 5.76                                              & \textbf{4.38 (0.15)}                                       \\
energy                            & 2.04  & 1.99                                                 & 1.38                                                & 0.9 (0.04)                                           & 1.62                                              & \textbf{0.61 (0.19)}                                       & 1.8  & 1.66                                                 & 2.09                                                 & 1.72 (0.08)                                           & 2.30                                              & \textbf{0.39 (0.02)}                                       \\
kin8nm                            & -0.9  & -0.95                                                & -1.2                                                & 6.89 (2.85)                                          & -1.14                                             & \textbf{-1.40 (0.01)}                                      & 0.1  & 0.1                                                  & 0.09                                                 & 0.26 (0)                                              & 0.09                                              & \textbf{0.06 (0.00)}                                       \\
navalpropulsion                   & -3.73 & -3.8                                                 & -5.63                                               & -3.03 (0.49)                                         & -5.73                                             & \textbf{-8.24 (0.01)}                                      & 0.01 & 0.01                                                 & \textbf{0}                                           & 0.01 (0)                                              & \textbf{0.00}                                     & \textbf{0.00 (0.00)}                                       \\
powerplant                        & 2.84  & 2.8                                                  & 2.79                                                & 5.24 (0.72)                                          & 2.83                                              & \textbf{2.66 (0.05)}                                       & 4.12 & 4.02                                                 & 4.11                                                 & 4.38 (0.02)                                           & 4.10                                              & \textbf{3.43 (0.08)}                                       \\
protein                           & 2.97  & 2.89                                                 & 2.83                                                & 21.12 (2.52)                                         & 3.12                                              & \textbf{2.48 (0.03)}                                       & 4.73 & 4.36                                                 & 4.71                                                 & 5.09 (0.01)                                           & 4.98                                              & \textbf{3.52 (0.02)}                                       \\
wine                              & 0.97  & \textbf{0.93}                                        & 0.94                                                & 1.92 (0.92)                                          & 1.15                                              & 1.00 (0.08)                                                & 0.64 & \textbf{0.62}                                        & 0.64                                                 & 0.73 (0.01)                                           & 0.65                                              & \textbf{0.62 (0.01)}                                       \\
yacht                             & 1.63  & 1.55                                                 & 1.18                                                & 0.48 (0.19)                                          & 0.76                                              & \textbf{-0.17 (0.11)}                                      & 1.02 & 1.11                                                 & 1.58                                                 & 1.86 (0.15)                                           & 1.00                                              & \textbf{0.44 (0.06)}                                       \\
yearprediction                    & 3.6   & 3.59                                                 & 3.35                                                & 7.44 (0.08)                                          & 3.58                                              & \textbf{3.22 (0.00)}                                       & 8.88 & 8.85                                                 & 8.89                                                 & 16.84 (0.08)                                          & 9.30                                              & \textbf{7.91 (0.04)}                                       \\ \hline
\textbf{Mean Rank}                & 4.9   & 3.4                                                  & 2.5                                                 & 4.7                                                  & 3.9                                               & \textbf{1.5}                                               & 3.7  & 2.6                                                  & 3.8                                                  & 4.6                                                   & 4                                                 & \textbf{1.3}                                               \\ \hline
\end{tabular}%
}
\caption{Results of the regression benchmark on 10 datasets.}
\label{tab:regression-benchmark}
\end{table*}

We analyze the impact of different search methods in AutoDEUQ on the uncertainty estimation. We compare the default AutoDEUQ (AgEBO) method (Fig.~1.a) with random search (RS-Mixed) (Fig.~2.a), AgE (AgE-Mixed) (Fig.~2.b), and BO (BO-Mixed) (Fig.~2.c). Note that RS, AgE, and BO do not consider the architecture and hyperparameter space separately. Instead, a configuration in the search space is given by a single vector of architecture decision variables and training hyperparameters.


We observe that the uncertainty estimates from the AutoDEUQ (RS-Mixed) are inferior to all other methods. 
AutoDEUQ (AgEBO) achieves more robust estimates than those of AutoDEUQ (AgE-Mixed) and AutoDEUQ (BO-Mixed). 
The estimates of epistemic uncertainty for AutoDEUQ (AgEBO), AutoDEUQ (AgE-Mixed), and AutoDEUQ (BO-Mixed) show a growing trend in the interpolation region as we move away from the training region. AutoDEUQ (BO-Mixed) has larger epistemic uncertainty in the interpolation region than AutoDEUQ (AgEBO) and AutoDEUQ (AgE-Mixed) have. 


The observed differences between the search methods can be attributed to the model diversity in the ensembles. To demonstrate this, we computed the architecture diversity for each method as follows. Each architecture was embedded as a vector of integers where each integer represents a choice for one of the decision variable of the neural architecture search space. To compute the diversity of an ensemble, we computed the pairwise Euclidean distance between the embeddings of the architectures composing the ensemble. Then, we kept only the upper triangle of the pairwise distance matrix (because it is symmetric) and normalized it by its norm.  We then computed the cumulative sum of the elements of this normalized triangular matrix, which gives us a scalar value representing diversity. AutoDEUQ (RS-Mixed) achieved the lowest diversity score (1.41), which also correlates with its poor epistemic uncertainty estimation. While AutoDEUQ (RS-Mixed) obtained diverse models for the catalog, they are not high-performing, and consequently the ensemble did not have diverse models. AutoDEUQ (AgE-Mixed) achieved a diversity score of 2.86, which resulted in a better epistemic uncertainty estimate in the interpolation region, but the estimates are poor in the extrapolation region. With a diversity score of 3.49, AutoDEUQ (BO-Mixed) obtained more diverse models, but they contributed to overly large epistemic uncertainty in the interpolation region and extrapolation regions. AutoDEUQ (AgEBO) achieved a diversity score of 3.17, which was in between that of AutoDEUQ (AgE-Mixed) and AutoDEUQ (BO-Mixed). Moreover, we found that the learning rate values obtained by AutoDEUQ (BO-Mixed) are more diverse than those obtained by AutoDEUQ (AgEBO). The training hyperparameter values obtained by these methods are given in the Appendix.

\subsection{Regression Benchmarks}

Here we compare our AutoDEUQ method with probabilistic backpropagation (PBP), Monte Carlo dropout (MC-Dropout), deep ensemble (Deep Ens.), distribution-free ensembles (DF-Ens.), and hyper ensemble (Hyper Ens.) methods. While PBP is selected as a candidate for Bayesian NN, MC-Dropout was selected for its popularity and simplicity. The Deep Ens. (with random initialization of weights, fixed architecture, and hyperparameters) will serve as a baseline method. The Hyper Ens. (ensemble with the same architecture but with different hyperparameters) is selected because it was a recently proposed high-performing ensemble method. 

 Ro assess the quality of uncertainty quantification methodologies, we used 10 regression benchmark datasets from the  literature~\cite{lakshminarayanan_simple_2017,hernandez-lobato_probabilistic_2015,gal_dropout_2016} (see the Appendix for a description of the datasets). We compare these methods using two metrics:  (1) negative log likelihood (NLL)  (i.e., how likely  the data is to be generated by the predicted normal distribution) and (2) root mean square error  (RMSE). These two metrics were widely adopted in the literature to compare the quality of uncertainty estimation. The metric values of PBP, MC-Dropout, Deep Ens., and DF-Ens. are copied from their corresponding papers~\cite{hernandez-lobato_probabilistic_2015,gal_dropout_2016,lakshminarayanan_simple_2017,pearce2018high}, respectively. Nevertheless, we extended and ran the Hyper Ens. method for regression based on the information provided in~\cite{wenzel_hyperparameter_2021}.

For each dataset, we ran AgEBO to generate a catalog of 500 models and used the \textit{greedy} selection strategy to construct ensembles of $K=5$ members. We repeated the experiments 10 times with different random seeds for the training/validation split and computed the mean score and its standard error. An exception was the \textit{yearprediction} dataset, which was run only 3 times because the dataset size was large.

The results are shown in Table~\ref{tab:regression-benchmark}. We  observe that AutoDEUQ obtains superior performance compared with the other methods with respect to both  NLL and RMSE. We computed the ranking of the methods for each dataset and computed the mean across the 10 datasets. This is shown in the last row of Table~\ref{tab:regression-benchmark}. AutoDEUQ with Greedy outperforms all of the other methods on 8 out of 10 datasets. On boston and wine, Hyper Ens. and MC Dropout have the lowest NLL and RMSE values. We note that, overall, the recently proposed Hyper Ens. performs worse than all the other methods. This performance can be attributed to the architecture used for regression in Hyper Ens., which is a simple multilayer perception network as described in the original paper~\cite{wenzel_hyperparameter_2021}. This further emphasizes the importance of and need for the architecture search for different datasets.

\section{Conclusion and Future Work}




We developed AutoDEUQ, an approach to automate the generation of deep ensembles for uncertainty quantification. 
We empirically demonstrated that epistemic uncertainty is best captured when the models considered in the ensemble are diverse (in hyperparameters and architecture), yet all the models perform  well and similarly on the validation set. This result is achieved by a two-step process: (1) using aging evolution and Bayesian optimization to jointly explore the neural architecture and hyperparameter space and generate a diverse catalog of models and (2) using greedy selection of models optimized with the negative log likelihood, to find models that are very different but all with high (and similar) performance. We conducted an extensive regression benchmark to compare AutoDEUQ with different classes of UQ methods, with and without ensembles. Our results confirm quantitatively what was observed on the toy example. The key ingredient of our technique is the diversity and predictive strength and homogeneity of the final ensemble.

Using a toy example, we performed an ablation study to visualize the impact of different components of AutoDEUQ on uncertainty estimation. This impact appears clearly in regions depleted in the  training samples. Compared with AutoDEUQ, methods optimizing either     hyperparameters independently   or architecture search underestimate epistemic uncertainty. 
Moreover, we conducted an extensive regression benchmark study to compare AutoDEUQ against different classes of UQ methods, with and without ensembles. Our results confirm quantitatively what was observed on the toy example. 

The key ingredient of our technique is the diversity and predictive strength and homogeneity of the final ensemble. AutoDEUQ is a computationally expensive method. However, the computational need can be controlled by restricting the search space and running model evaluations in parallel.



Our future work will include (1) applying AutoDEUQ on larger datasets to assess its scalability, (2) evaluating AutoDEUQ on a classification benchmark, and (3) seeking theoretical insights into the quality of epistemic uncertainty under the various data generation assumptions.
\nocite{hernandez-lobato_probabilistic_2015}
\nocite{hansen_neural_1990}
\nocite{lakshminarayanan_simple_2017}
\nocite{wenzel_hyperparameter_2021}

\footnotesize{
\section{Acknowledgement}
This work was supported by the U.S.\ Department of Energy, Office of Science, Advanced Scientific Computing Research, under Contract DE-AC02-06CH11357 and DOE Early Career Research Program award. We are grateful for the use of the computing resources in the Joint Laboratory for System Evaluation  and Leadership Computing Facility at Argonne.
}
\clearpage
\clearpage

\section{Appendix A}

\begin{algorithm2e}[!ht]
\footnotesize
\DontPrintSemicolon
\SetInd{0.5em}{0.5em}
\SetAlgoLined
\SetKwInOut{Input}{inputs}\SetKwInOut{Output}{output}
\SetKwFunction{RandomPoint}{random\_sample}
\SetKwFunction{SubmitEval}{submit\_for\_training}
\SetKwFunction{GetFinishedEval}{check\_finished\_training}
\SetKwFunction{Push}{push}
\SetKwFunction{EmptyList}{EmptyList}
\SetKwFunction{RandomSample}{random\_sample}
\SetKwFunction{SelectParent}{select\_parent}
\SetKwFunction{Mutate}{mutate}
\SetKwFunction{Tell}{tell}
\SetKwFunction{Ask}{ask}

\SetKwFor{For}{for}{do}{end}
\Input{P: population size, S: sample size, W: workers}
\Output{$\mathcal{E}$: ensemble of models}
    {\color{orange} \tcc{Initialization for AgEBO}}
    $population \leftarrow$ create\_queue($P$) \tcp{Alloc empty Q of size P}
    $BO \leftarrow$ Bayesian\_Optimizer()\\
    \For{$i\leftarrow 1$ \KwTo $W$}{
        $config.\theta_a \leftarrow$ \RandomPoint{$\Theta_a$}\\
        $config.\theta_h \leftarrow$ \RandomPoint{$\Theta_h$}\\
        \SubmitEval{config} \tcp{Nonblocking}
    }
    
    {\color{orange} \tcc{Optimization loop for AgEBO}}
    \While{stopping criterion not met}{
        \tcp{Query results}
        $results \leftarrow$ \GetFinishedEval()\\
        $\mathcal{C} \leftarrow \mathcal{C} \cup results$ \tcp{Add to catalogue population}
        \If{$|results| > 0$}{
            $population.$\Push{results} \tcp{Aging population}
            \tcp{Generate hyperparameter configs}
            $BO.$\Tell{$results.\theta_h, results.valid\_score$}\\
            $next \leftarrow$ $BO.$\Ask{$|$results$|$} \tcp{Generate architecture configs}
            \For{$i\leftarrow 1$ \KwTo $|results|$}{
                \eIf{$|population| = P$}{
                    $parent.config \leftarrow$ \SelectParent{population,S}\\
                    $child.config.\theta_a \leftarrow$ \Mutate{$parent.\theta_a$}
                }
                {
                $child.config.\theta_a \leftarrow$ \RandomPoint{$\Theta_a$}
                }
                $child.config.\theta_h \leftarrow next[i].\theta_h$ \\
                \SubmitEval{$child.config$} \tcp{Nonblocking}
            }
        }
    }
    {\color{orange} \tcc{Initialization for ensemble construction}}
    $\mathcal{E} \leftarrow \{\}$\\
    $min\_loss \leftarrow +\infty$\\
    {\color{orange} \tcc{Model selection}}
    \While{$|\mathcal{E}.unique()| \leq K$}{
        $\theta^* \leftarrow \argmin_{\theta \in \mathcal{C}} \ell(\mathcal{E} \cup \{ \theta \},X,y)$ \\
        \eIf{$\ell(\mathcal{E} \cup \{ \theta^* \},X,y) \leq min\_loss$}{
            $\mathcal{E} \leftarrow \mathcal{E} \cup \{ \theta^* \}$ \\
            $min\_loss \leftarrow \ell(\mathcal{E},X,y)$
        }{
            return $\mathcal{E}$
        }
    }
    return $\mathcal{E}$
 \caption{AutoDEUQ for ensemble construction}
 \label{alg:AgEBO}
\end{algorithm2e}

\subsection{Experimental Settings}

\normalsize

We conducted our experiments on the ThetaGPU system at the Argonne Leadership Computing Facility. ThetaGPU is composed of 24 nodes, each composed of 8 NVIDIA A100 GPUs and 2 AMD Rome 64-core CPUs. 

For the {\bf generation of a catalog of models} we use different allocations (i.e., number of nodes) depending on the dataset size. During the search, 1 process only using the CPU is allocated for the search algorithm; then neural network configurations (hyperparameters and architecture) are sent to parallel workers for the training. Each worker corresponds to a single GPU. Therefore,  1 node had 8 parallel workers. 
For the {\bf construction of an ensemble}, we load all checkpointed models on different GPU instances to perform parallel inferences and then save the predictions to apply the greedy strategy. 

On the software side, we used Python 3.8.5.  The core of our dependencies is composed of TensorFlow 2.5.0, TensorFlow-Probability 0.13.0, Ray 1.4.0, Scikit-Learn 0.24.2, and Scipy 1.7.0.

\subsection{Neural Architecture Search}

\begin{figure}[!h]
    \centering
    \includegraphics[width=0.7\linewidth]{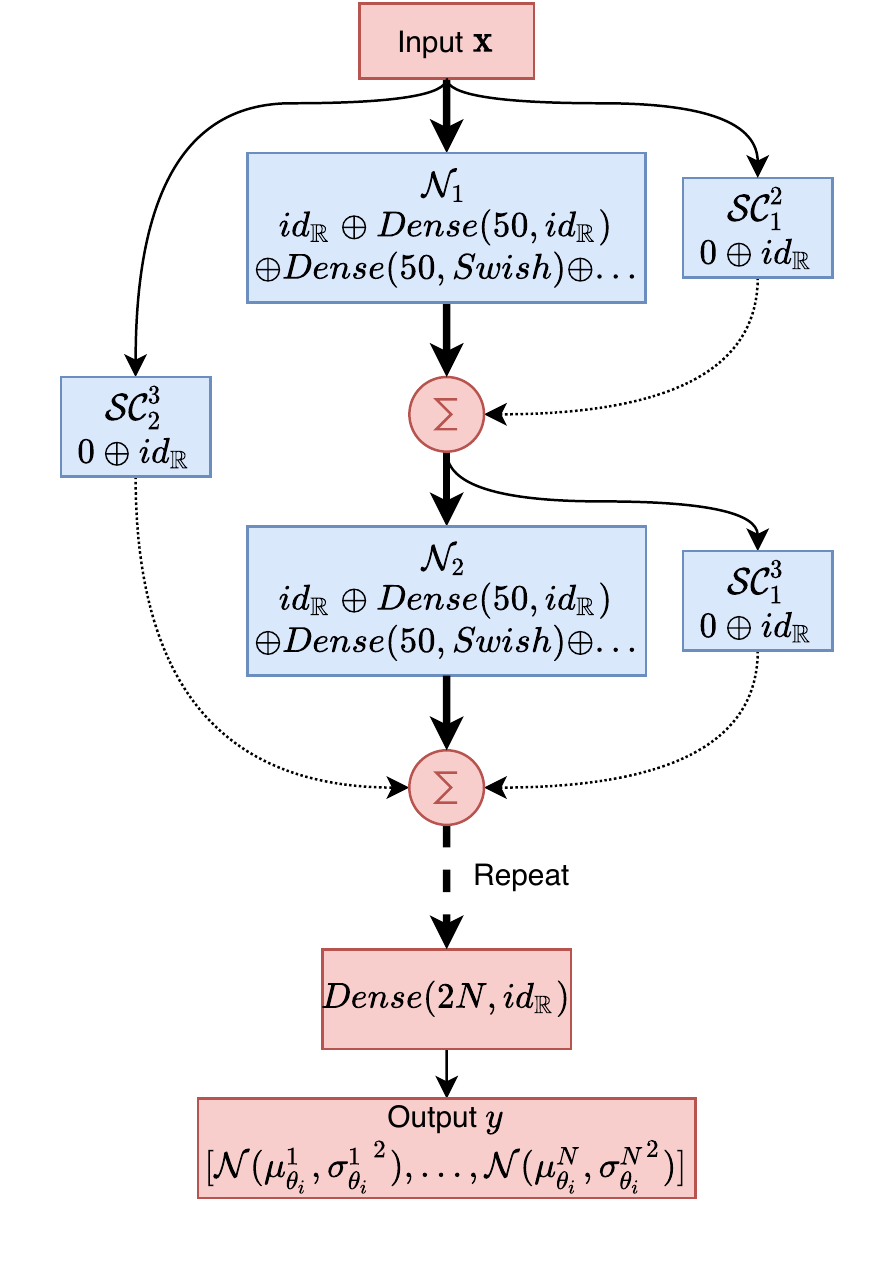}
    \caption{Search space of fully connected neural networks with regression outputs}
    \label{fig:search_space_mlp}
\end{figure}

In AutoDEUQ, we used a neural architecture search space of fully connected  neural networks with possible skip connections. A visualization of this search space is presented in Figure~\ref{fig:search_space_mlp}. $N$ denotes the number of output variables.

\subsection{Benchmark Datasets}
In Table~\ref{tab:regression_benchmark_datasets_details} we give details about the different datasets used in our regression benchmark. These datasets are from the UCI Machine Learning Repository~\cite{Dua:2019}.
\begin{table}[!h]
\centering
\resizebox{0.8\textwidth}{!}{%
\begin{tabular}{l|rr}
Dataset's Name  & \multicolumn{1}{l}{Number of Samples} & \multicolumn{1}{l}{Feature Size} \\ \hline
boston          & \multicolumn{1}{r|}{506}              & 13                               \\
concrete        & \multicolumn{1}{r|}{1030}             & 8                                \\
energy          & \multicolumn{1}{r|}{768}              & 8                                \\
kin8nm          & \multicolumn{1}{r|}{8192}             & 8                                \\
navalpropulsion & \multicolumn{1}{r|}{11934}            & 16                               \\
powerplant      & \multicolumn{1}{r|}{9568}             & 4                                \\
protein         & \multicolumn{1}{r|}{45730}            & 9                                \\
wine            & \multicolumn{1}{r|}{1599}             & 11                               \\
yacht           & \multicolumn{1}{r|}{308}              & 6                                \\
yearprediction  & \multicolumn{1}{r|}{515345}           & 90                              
\end{tabular}%
}
\caption{Description of the different datasets used in the regression benchmark.}
\label{tab:regression_benchmark_datasets_details}
\end{table}

\bibliographystyle{IEEEtran}
\bibliography{main.bib}


\end{document}